\def\year{2020}\relax
\documentclass[letterpaper]{article} % DO NOT CHANGE THIS
\usepackage{aaai20}  % DO NOT CHANGE THIS
\usepackage{times}  % DO NOT CHANGE THIS
\usepackage{helvet} % DO NOT CHANGE THIS
\usepackage{courier}  % DO NOT CHANGE THIS
\usepackage[hyphens]{url}  % DO NOT CHANGE THIS
\usepackage{graphicx} % DO NOT CHANGE THIS
\usepackage{graphicx}  % DO NOT CHANGE THIS
\usepackage{subfig}
\usepackage{lineno,hyperref}
\usepackage{amsfonts}
\usepackage{multirow}
\usepackage{bm}
\modulolinenumbers[5]
\usepackage{rotating}
\usepackage{booktabs}

\usepackage{makecell}

\usepackage{amsmath}
\usepackage{amsfonts}
\usepackage{algorithm}
\usepackage{algorithmic}

\title{Efficient Novelty-Driven Neural Architecture Search}
\author{Miao Zhang\textsuperscript{\rm 1,2}, Huiqi Li\textsuperscript{\rm 1}, Shirui Pan\textsuperscript{\rm 3}, Taoping Liu\textsuperscript{\rm 2}, Steven Su\textsuperscript{\rm 2}\\ 
	\textsuperscript{\rm 1}School of Information and Electronics, Beijing Institute of Technology, China\\
	 \textsuperscript{\rm 2}Faculty of Engineering and Information Technology, University of Technology Sydney, Australia\\
	\textsuperscript{\rm 3}Faculty of Information Technology, Monash University, Australia\\
	miao.zhang-2@student.uts.edu.au,huiqili@bit.edu.cn, shirui.pan@monash.edu, taoping.liu@student.uts.edu.au, steven.su@uts.edu.au  % email address must be in roman text type, not monospace or sans serif
}

\begin{document}

\maketitle

\begin{abstract}
One-Shot Neural architecture search (NAS) attracts broad attention recently due to its capacity to reduce the computational hours through weight sharing. However, extensive experiments on several recent works show that there is no positive correlation between the validation accuracy with inherited weights from the supernet and the test accuracy after re-training for One-Shot NAS. Different from devising a controller to find the best performing architecture with inherited weights, this paper focuses on how to sample architectures to train the supernet to make it more predictive. A single-path supernet is adopted, where only a small part of weights are optimized in each step, to reduce the memory demand greatly. Furthermore, we abandon devising complicated reward based architecture sampling controller, and sample architectures to train supernet based on novelty search. An efficient novelty search method for NAS is devised in this paper, and extensive experiments demonstrate the effectiveness and efficiency of our novelty search based architecture sampling method. The best architecture obtained by our algorithm with the same search space achieves the state-of-the-art test error rate of 2.51\% on CIFAR-10 with only 7.5 hours search time in a single GPU, and a validation perplexity of 60.02 and a test perplexity of 57.36 on PTB. We also transfer these search cell structures to larger datasets ImageNet and WikiText-2, respectively.
\end{abstract}

\section{Introduction}
Neural architecture search (NAS) recently attracts massive interests from deep learning community since it could relieve experts from a labor-intensive and time-consuming neural network design process \cite{zoph2016neural,elsken2018neural,liu2017hierarchical}. Despite its capacity to find competitive architectures, NAS is  computationally expensive. Zoph et al. \shortcite{zoph2018learning} spends more than 1800 GPU days based on reinforcement learning (RL) and Real et al. \shortcite{real2018regularized} uses 450 GPUs for 7 days through evolutionary algorithm (EA) to train the model. To improve the efficiency of NAS, several works have been proposed, including performance prediction  \cite{baker2017accelerating}, weight generation \cite{brock2017smash,zhang2018graph}, and also the popular weight sharing method \cite{pham2018efficient}. 

Weight sharing, also called One-Shot NAS \cite{pham2018efficient,bender2018understanding},  defines a supernet subsuming all possible architectures in the search space, where those architectures directly inherit weights from the supernet to avoid training from scratch. ENAS \cite{pham2018efficient}  utilizes the validation accuracy with shared weights as the reward to optimize the architecture sampling policy in a RL method.  Following up works \cite{liu2018darts,luo2018neural} 
relax architectures into continuous space and optimize the architecture with respect to its validation accuracy with inherited weights through gradient descent. As the architectures are measured by being associated with inhered weights from the supernet, an important assumption in the weight-sharing NAS is that the measurement of architectures with inhered weights approximates to fully trained architectures, or at least be highly predictive. However, several recent works \cite{bender2018understanding,singh2019study,sciuto2019evaluating} point out that there is no positive correlation between the validation accuracy with inherited weights from the supernet and the test accuracy after re-training for these One-Shot NAS methods. This indicates that we could not utilize the validation accuracy with inherited weights as useful feedback for controller improvement. In other words, searching for the optimal architecture for DNN based on weight sharing is deceptive because architectures with optimal performance on proxy task are not guaranteed to perform best in the target task \cite{cai2018proxylessnas}.

As the validation accuracy with inherited weights is deceptive \cite{sciuto2019evaluating,singh2019study}, solely optimizing for this deceptive reward without encouraging intelligent exploration usually leads to local optima. Different from RL controller or gradient method, novelty search is potentially able to alleviate this problem by encouraging the agent to visit unexplored areas rather than those areas with high performance. As suggested by curiosity-driven exploration in deep reinforcement learning \cite{pathak2017curiosity,conti2018improving}, novelty-seeking could help the agent to learn new knowledge and avoid local optima in RL domains with deceptive or sparse rewards. Instead of devising a complicated controller, we innovatively introduce novelty search to NAS, which samples architectures to train supernet through novelty search to make the supernet more predictive. A weight-sharing based single-path model is adopted to reduce computational cost and memory demand, where all candidate architectures share weights and only the weights in a single-path architecture are optimized in each step. Our approach samples the architecture that is most different from previously visited architectures to train the supernet, and only the shared weights of the sampled architecture and supernet are optimized in the training procedure.  Our contributions are summarized as follows.

\begin{itemize}
\item Firstly, a novelty based search mechanism is innovatively applied to NAS for sampling architectures to train supernet, and an efficient approach is devised to sample architectures with novelty. 
\item Secondly, we apply a weight-sharing based single-path model to neural architecture search, which could reduce not only the computational cost but also the memory demand greatly.

\item Thirdly, extensive experimental results illustrate the superiority of our method which achieves remarkable performance on CIFAR10 and PTB with efficiency. Our approach obtains the state-of-the-art test error of 2.51\% for CIFAR10 with only 7.5 hours search time in a single GPU, and a competitive validation perplexity of 60.27 and a test perplexity of 57.8 on PTB with 4 hours search time, and achieves a validation perplexity of 60.02 and a test perplexity of 57.36 on PTB when combined with performance reward function. We also transfer these search cell structures to larger datasets ImageNet and WT2. Experimental datasets and source codes could be found in supplemental material \footnote{Experimental datasets and source codes could be found in the supplemental material. All codes and datasets will be releases after acceptance.}.
\end{itemize}

\section{Background}

\subsection{Neural Architecture Search}

Neural architecture search (NAS) recently has attracted increasing attention to automatically design neural architecture to relieve human experts from the labor-intensive and time-consuming neural network design process. The search space of neural architecture $\mathcal{A}$ is generally represented as a directed acyclic graph (DAG), and the subgragh in the search space is denoted as $\alpha\in \mathcal{A}$ corresponding to a neural architecture $\mathcal{U}(\alpha,w)$ with weights $w$. NAS aims to find a subgraph $\alpha$ with best validation loss after being trained on the training set, as
\begin{equation} \label{[1]}
\alpha^* =\underset{\alpha \in \mathcal{A}}{\mathtt{argmin}}\ \mathcal{L}_{\mathtt{val}}( \mathcal{U}(\alpha,w_\alpha))
\end{equation}
where $\mathcal{L}_{\mathtt{val}}$ is the loss function on the validation set, and $w_\alpha$ are the weights of the architecture after trained on the training set to minimize the training loss $\mathcal{L}_{train}$:
\begin{equation} \label{[2]}
w_\alpha =\underset{w}{\mathtt{argmin}}\ \mathcal{L}_{\mathtt{train}}( \mathcal{U}(\alpha,w))
\end{equation}

Early NAS works adopt a nested manner to optimize weights and architectures, which samples numerous architectures to be trained on the training set and utilize EA \cite{real2018regularized} or RL \cite{zoph2016neural} to find promising architectures based on those evaluated architectures. Guo et al. \shortcite{guo2018irlas} further propose an inverse reinforcement learning method to force the agent to search for architectures that are similar to human-designed networks. These approaches have a high computational demand because evaluating an architecture is computationally expensive, which makes this straightforward manner inefficient and unaffordable, and a lot of NAS approaches are motivated by reducing computational cost \cite{zoph2018learning,liu2018progressive,baker2017accelerating,baker2017accelerating,brock2017smash,zhang2018graph}. 

Recently, a weight sharing mechanism (also called as One-Shot) is adopted in NAS \cite{pham2018efficient,liu2018darts}, which could greatly reduce the search time to less than 1 GPU day. Instead of training separate architectures, weight sharing strategy encodes the whole search space $\mathcal{A}$ as a supernet $\mathcal{U}(\mathcal{A},\mathcal{W})$, and all candidate architectures $\mathcal{U}(\alpha,w)$ directly inherit weights from the weights $\mathcal{W}$ of supernet. Only the supernet is trained in the architecture search phase for weight sharing NAS approaches, so it is able to reduce the time for architecture search greatly. The weight sharing based NAS contain two sequential steps 1) the supernet training:
\begin{equation} \label{[3]}
\mathcal{W}_{\mathcal{A}}=\underset{\mathcal{W}}{\mathtt{argmin}}\ \mathcal{L}_{\mathtt{train}}( \mathcal{U}(\mathcal{A},\mathcal{W}))
\end{equation}
and 2) architecture selection:
\begin{equation} \label{[4]}
\alpha^* =\underset{\alpha \in \mathcal{A}}{\mathtt{argmin}}\ \mathcal{L}_{\mathtt{val}}( \mathcal{U}(\alpha,\mathcal{W}_{\mathcal{A}}(\alpha)))
\end{equation}

The key for weight-sharing based NAS is how to sample architectures for supernet training to make the inhered weights $\mathcal{W}_{\mathcal{A}}(\alpha)$ approximate to the fully trained weights $w_\alpha$ or be highly predictive, where ENAS \cite{pham2018efficient} utilizes an LSTM controller to sample architectures. Recent weight sharing approaches relax architectures in to continuous space $\mathcal{A}_{\theta}$ \cite{liu2018darts,GDAS,xie2018snas,wu2018fbnet,zhou2019bayesnas,luo2018neural}, where $\alpha_\theta$ is continue parameters representing architectures, and utilize the gradient descent or stochastic methods to optimize weights and architectures, as:

\begin{equation} \label{[5]}
(\alpha_\theta^*, \mathcal{W}_{\alpha_\theta^*})=\underset{\alpha_\theta,\mathcal{W}}{\mathtt{argmin}}\ \mathcal{L}_{\mathtt{train}}( \mathcal{U}(\mathcal{A}_\theta,\mathcal{W}))
\end{equation}

Although gradient methods or stochastic methods in continuous space make the architecture search much more efficient, it has much higher memory requirements that it needs to train whole weights in the supernet. ProxylessNAS \cite{cai2018proxylessnas} further utilizes binary gates to zero out real-valued architecture parameters and only one path of the supernet is activated during supernet training, thus reduces the memory requirement to the same level as training a single architecture. It achieves remarkable test accuracy in CIFAR10 and ImageNet, while introduces one more controller and makes the architecture search phase more complicated. Different from searching in the continuous space, Casale et al. \shortcite{casale2019probabilistic} propose a probabilistic approach PARSEC to sample architectures without continuous relaxation, where it uses an Importance-Weighted Monte Carlo empirical Bayes to define the architecture distribution.

Extensive experimental analysis in recent works \cite{bender2018understanding} demonstrates that it is possible to efficiently sample architectures for supernet training without any complex controllers for NAS, and Guo et al. \shortcite{guo2019single} and Li et al. \shortcite{li2019random} respectively utilize the simple uniform sampling and random sampling method as the architecture search controller to sample architectures for supernet training. The weight sharing is adopted in both of them to reduce the computational cost, and the memory requirements are all same as training a single architecture that only one path of the supernet is activated in each step of the architecture search phase.

\subsection{Novelty Search}

Novelty search comes from the evolutionary community \cite{lehman2011abandoning,real2018regularized}, which encourages the population to search for notably different areas to enhance the exploration. This suggested approach utilizes the novelty as the stepping stone instead of the reward function, which makes it easy to get out of local optima in return. Previous novelty search based evolutionary algorithms \cite{stanley2002evolving,lehman2011abandoning} had shown their superiority in searching for small neural networks, and recent works on deep reinforcement learning \cite{conti2018improving} also suggested that hybridized with novelty search, evolutionary algorithm could effectively avoid local optima in RL domains with deceptive reward functions. We investigate the effects of novelty search on neural architecture search in this paper, where we detailed present how to use the novelty search mechanism as the controller to sample architectures for training the supernet in the following section.

\begin{figure*}
 \subfloat[Search space for CNN]{
  \begin{minipage}{6cm}
      \includegraphics[width=6cm,height=3.5cm]{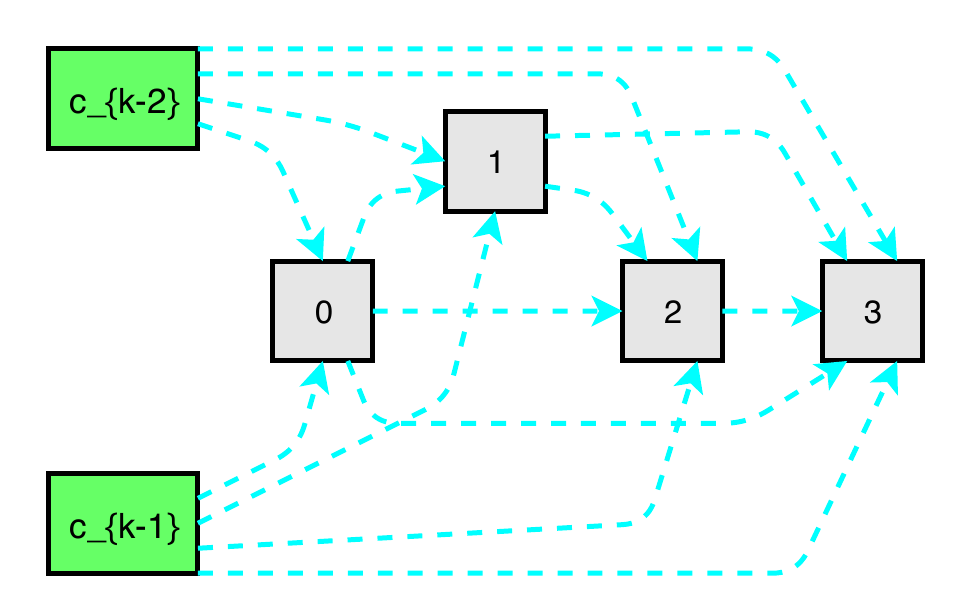}
  \end{minipage}
 }
  \subfloat[Normal cell learned on CIFAR-10]{
  \begin{minipage}{6cm}
      \includegraphics[width=6cm,height=3.5cm]{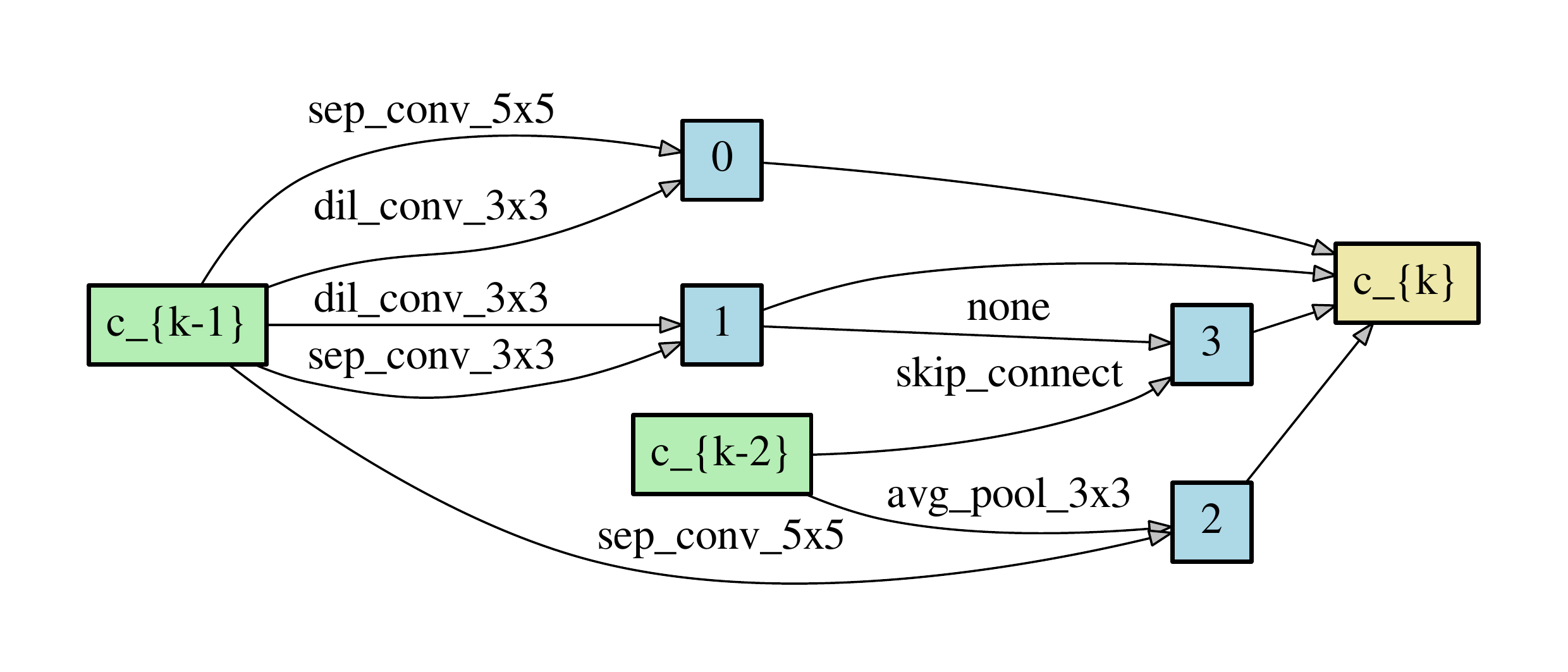}
  \end{minipage} 
  }
   \subfloat[Reduction cell learned on CIFAR-10]{
  \begin{minipage}{6cm}
      \includegraphics[width=6cm,height=3.5cm]{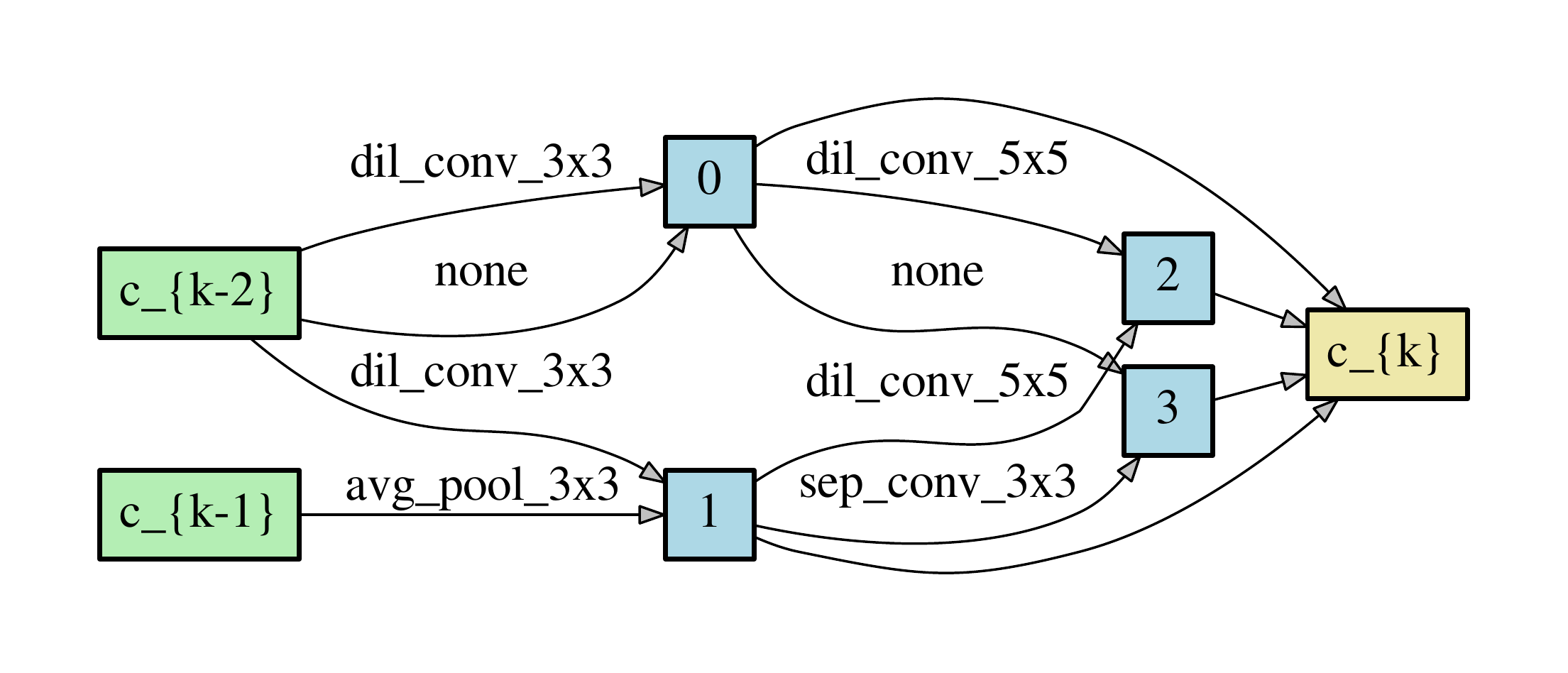}
  \end{minipage}
  }

    \subfloat[Search space for RNN]{
  \begin{minipage}{6cm}
      \includegraphics[width=6cm,height=3.5cm]{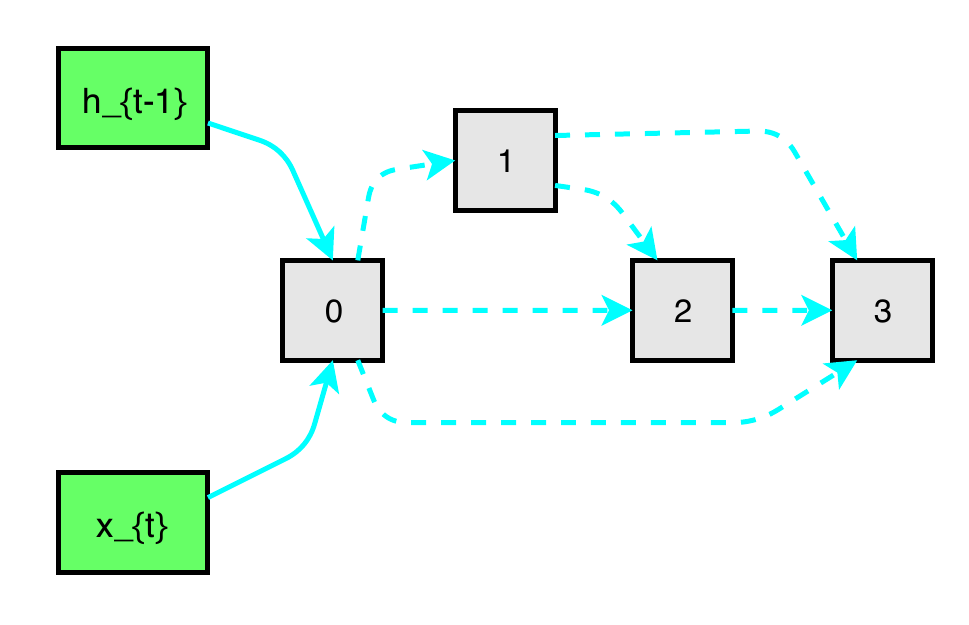}
  \end{minipage} 
  }
   \subfloat[Recurrent cell learned on PTB]{
  \begin{minipage}{6cm}
      \includegraphics[width=12cm,height=3.5cm]{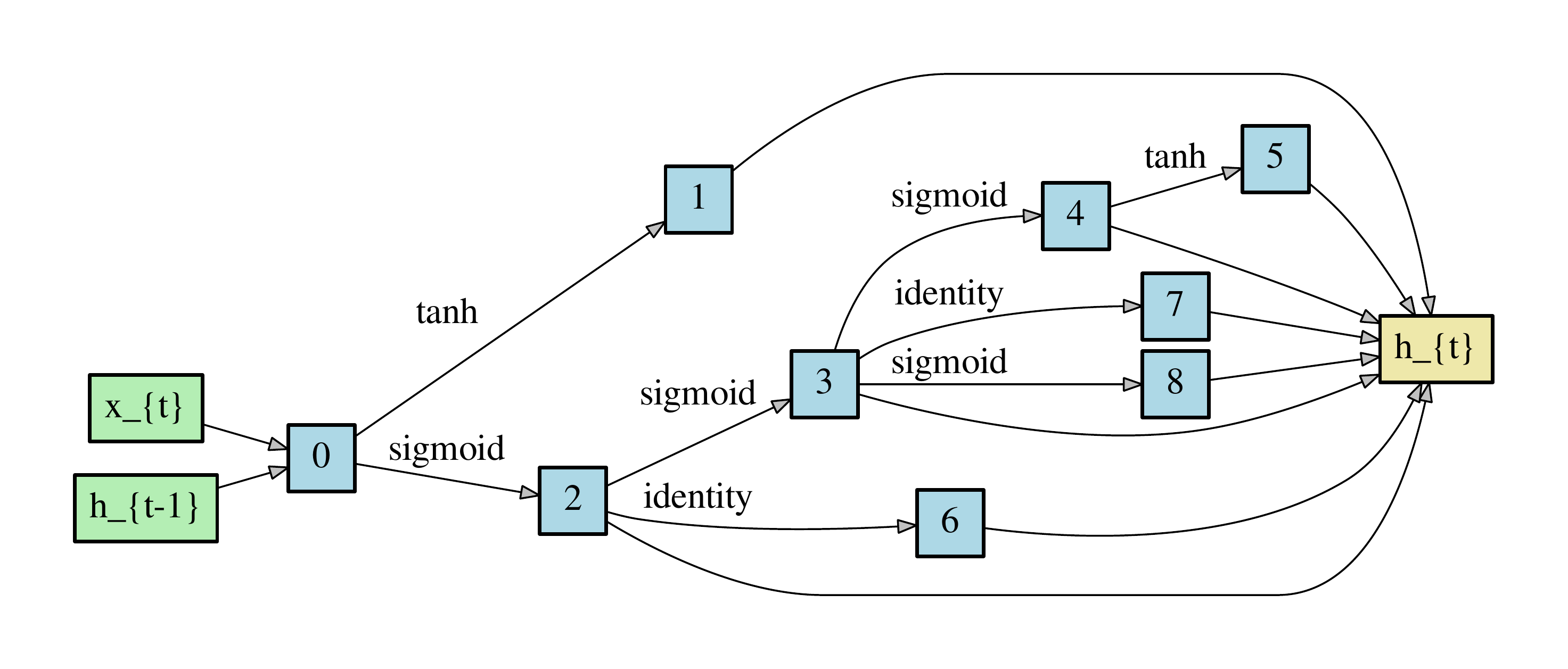}
  \end{minipage}
  }
  \caption{Search space and best cell structures found by our algorithm. For CNN cell, each node needs to select two former nodes with applied operations as its input. As to RNN cell, each node only needs to select one former node with applied operation as its input. (d) is the search space of RNN with only $B=3$ nodes as an example which should be 8 in this paper. The outputs for the three types of cells are the summation of outputs for all nodes in the cells.}
 \label{figure1}
\end{figure*}

\section{Methodology}
In this section, we will describe our efficient novelty-driven neural architecture search (EN$^2$AS). The framework of our approach is similar to Random Search WS \cite{li2019random} that weight sharing is adopted and only one-path is activated in each step of supernet training. Differently, our approach adopts a novelty based mechanism to search for the promising architectures during the architecture search phase, which could effectively avoid local optima and enhance the exploration. Algorithm~\ref{alg:algorithm1} presents a simple implementation of EN$^2$AS, and we detailed describe the search space, architecture sampling for supernet training based on novelty search and also discuss two approaches for architecture selection from trained supernet in the following section.

\begin{algorithm}[tb]
\caption{EN$^2$AS}
\label{alg:algorithm1}
\textbf{Input}: Training datase $\mathbb{D}_{train}$, validation dataset $\mathbb{D}_{val}$, test dataset $\mathbb{D}_{test}$, randomly initialized $W$, initial architecture archive $A=\emptyset$, maximum number of stored architectures $S$, batch size $b$, training iteration $T$
\begin{algorithmic}[1]
\FOR{$i=1,2,...,(T*\mathtt{size}(\mathbb{D}_{train})/b)$}
\IF{{$\mathtt{size}(A)<S$}}
\STATE randomly sample an architecture $\alpha$, and update the shared weights $\mathcal{W}_{A}(\alpha)$ by descending $\nabla_{\mathcal{\mathcal{W}}_{A}(\alpha)}\mathcal{L}_{train}(\mathcal{W}_{A}(\alpha))$
\STATE add architecture $\alpha$ into $A$
\ELSE
\STATE randomly select an architecture $\alpha_\theta^{m}$ from $A$, update it according Eq.\eqref{[8]} or Eq.\eqref{[9]}, and replace $\alpha_\theta^{m}$ with $\alpha_\theta^{m'}$
\STATE apply round operation on the updated architecture to obtain $\alpha$, and update the shared weights $\mathcal{W}_{A}(\alpha)$ by descending $\nabla_{\mathcal{W}_{A}(\alpha)}\mathcal{L}_{train}(\mathcal{W}_{A}(\alpha))$
\ENDIF
\ENDFOR
\STATE Perform random search or evolutionary algorithm on the trained supernet with validation dataset $\mathbb{D}_{val}$ to get $\alpha^*$ based on Eq.\eqref{[4]}
\STATE Retrain the most promising architecture with enough training iterations, and get the best performance on the test dataset
\end{algorithmic}
\textbf{Return}: architecture $\alpha^*$ with best performance

\end{algorithm}

\subsection{Search space}
The search space design plays an important role in NAS\cite{elsken2018neural}, and we consider a common search space used by \cite{real2018regularized,xie2018snas,liu2018darts,li2019random} for fair comparison. Architectures in the search space are represented as directed acyclic graphs (DAG), and we search for a computation cell that could be stacked to form the final architecture. The cell in our CNN is represented as a DAG with $B$ ordered nodes, where each node contains two inputs which are previous notes' outputs after being applied operation. The inputs of the cell are the outputs of two of its former cells' outputs, and its outputs is the summation of all nodes' outputs. We include 7 types of operations and also the ``zero" operation in our CNN architecture as shown in Fig.\ref{[1]}(a)(b)(c), and set $B=4$ for all cells. There are two types of cells in our CNN: normal cell and reduction cell, encode as $(\alpha_{normal},\alpha_{reduce})$, and reduction cells only locate in the 1/3 and 2/3 of the total depth of our network. 

The architecture search space for RNN is also the same as \cite{liu2018darts,li2019random}, where each cell contains $B=8$ ordered nodes, and the input of each node is the output of one of its previous nodes after being applied operation. The hidden state $h_t$ is calculated based on the input $i_t$ and its former hidden state $h_{t-1}$, and the output of the cell is the summation of outputs of all nodes. We also include 4 types of operations in our RNN architecture as shown in Fig.\ref{[1]}(d)(e), and we need only encode one type cell in RNN architecture. All setups of search space for CNN and RNN could also be found in \cite{liu2018darts,li2019random}.

\subsection{Single Path Supernet Training based on Novelty Search}
As described in Eq.\eqref{[3]}, the inherited weights $\mathcal{W}_{\mathcal{A}}(\alpha)$ of architecture $\alpha$ from the supernet $\mathcal{A}$ should approximate to the optimal weights $w_\alpha$ or be highly predictive. Therefore, the key to weight sharing based NAS is how to train the supernet. As discussed in \cite{li2019random,bender2018understanding}, a complicated reward gradient-based architecture sampling controller maybe not necessary for One-Shot NAS, and a random or uniform architecture sampling method could also achieve competitive results. Recent work \cite{conti2018improving} on Deep Reinforcement learning demonstrates the effectiveness of novelty search that it could help the agent get out of local optimal when the reward function is very deceptive. In this paper, we utilize the novelty search to sample architectures for supernet training in One-Shot NAS.

The novelty search policy is defined as $\pi$ and a behavior characterization $b(\pi)$ is to describe its behavior. During the architecture search phase, every architecture $\alpha$ sampled from $\pi$ is described as $b(\pi_{\alpha})$ and added into archive $A$ after calculating the novelty particular policy $N(b(\pi_{\alpha}),A)$. A simple and common novelty measurement is to calculate the mean distance of $\alpha$ and its k-nearest neighbors from $A$:

\begin{equation} \label{[6]}
\resizebox{.9\linewidth}{!}{$
    \displaystyle
\begin{aligned}
& N(\alpha,A)=N(b(\pi_{\alpha}),A)=\frac{1}{\left | S \right |}\sum_{j\in S}\left \|b(\pi_{\alpha})-b(\pi_j)  \right \|_2\\
& S=kNN(b(\pi_{\alpha}),A)={b(\pi_1),b(\pi_2),...,b(\pi_k)}
\end{aligned}
$}
\end{equation}

However, the distance calculation between neural architectures is not efficient because we need to compare all nodes and connections of two subgraphs, and calculating distances between the sampled architecture and all previously visited architectures in every search step is unrealistic. In this section, we introduce an archive based novelty search to relieve the high computational complexity for the novelty calculation. Given an archive $A_\theta$ containing a fixed number of continuous parameters representation of sampled architectures as $\alpha_{\theta}^{i}=\alpha_{\theta}+\sigma\epsilon_i$, the gradient of expected novelty could be approximated as \cite{conti2018improving}:
\begin{equation} \label{[7]}
\resizebox{.9\linewidth}{!}{$
    \displaystyle
\begin{aligned}
\nabla_{\alpha_\theta}\mathbb{E}_{\epsilon \sim \mathcal{N}(0,I)}[N(\alpha_\theta+\sigma \epsilon ,A)|A]\approx \frac{1}{n\sigma}\sum_{i=1}^{n}N(\alpha_\theta^i,A)\epsilon_i
\end{aligned}
$}
\end{equation}
where $\epsilon_i \sim \mathcal{N}(0,I)$, $\alpha_\theta^t$ is the \textsl{i-th} architecture with continuous parameters representation in the archive, $n$ is the number of sampled perturbations to $\alpha_\theta^t$, and the archive is fixed at the beginning of the iteration and updated at the end. Eq. \eqref{[7]} demonstrates how to change the current architectures could increase the novelty of the archive, and we could update \textsl{m-th} architecture in the archive according:
\begin{equation} \label{[8]}
\alpha_\theta^{m'}\leftarrow \alpha_\theta^{m}+\gamma \frac{1}{n\sigma}\sum_{i=1}^{n}N(\alpha_\theta^{m,i},A)\epsilon_i
\end{equation}
where $\gamma$ is the stepsize. In this way, we only need to calculate the distance of the sampled architecture and an archive with a fixed number of architectures in every search step. It is straightforward to randomly select an architecture from the archive, and update it accordingly to optimize novelty. In our practical implementation, only the architectures store in the archive are continuous, and they are also applied with the round operation before calculating the distance between sampled architectures and them.

Suggested by Conti et al. \shortcite{conti2018improving}, combining the performance reward and novelty could help the agent to not only avoid local optimal but also guide to search for better areas, this paper also tries to add the gradient of expected performance reward into adjusting the current architectures. Similar to Eq.\eqref{[8]}, we could update \textsl{m-th} architecture in the archive according:
\begin{equation} \label{[9]}
\resizebox{.9\linewidth}{!}{$
    \displaystyle
\alpha_\theta^{m'}\leftarrow \alpha_\theta^{m}+\gamma \frac{1}{n\sigma}\sum_{i=1}^{n}w\cdot ACC(\mathcal{W}_{\mathcal{A}}(\alpha_\theta^{m,i}))\epsilon_i+(1-w)\cdot N(\alpha_\theta^{m,i},A)\epsilon_i
$}
\end{equation}
where $ACC(\mathcal{W}_{\mathcal{A}}(\alpha))$ is the validation accuracy of $\alpha$ with inhered weights from the supernet, $w$ is a trade-off between the performance reward and novelty, which is defaulted set as 0.5 in this paper.

\subsection{Model Selection}
Because only inference occurs in the architecture selection from trained supernet, it is possible to sample enough architectures to find the most promising architecture based on Eq.\eqref{[4]}, where random search and evolutionary algorithms are the two most common methods \cite{li2019random,guo2019single,brock2017smash} to solve it. Random search is a simple but competitive method for architecture selection, which randomly samples numerous architectures to find the most promising one. Since evaluating an architecture is very efficient based on the trained supernet, it is possible to utilize a heuristic approach to find the best architecture. Guo et al. \shortcite{guo2019single} utilizes a baseline evolutionary algorithm for architecture selection from the trained supernet, which shows its superiority than random search.

In this paper, we adopt the validation accuracy as the optimizing goal in modeling selection as:
\begin{equation} \label{[10]}
\underset{\alpha}{\mathtt{maximize}}\quad ACC(\mathcal{W}_{\mathcal{A}}(\alpha))
\end{equation}
where $ACC(\mathcal{W}_{\mathcal{A}}(\alpha))$ is the validation accuracy of $\alpha$ with inhered weights from the supernet.

\begin{table*}\footnotesize 
\centering
\begin{tabular}
{lcccccc}
\toprule
\makecell[c]{Method}&\makecell[c]{Test Error\\(\%)}&\makecell[c]{Parameters\\(M)}&\makecell[c]{Search Cost\\(GPU Days)}&\makecell[c]{Memory\\Consumption}&\makecell[c]{Search\\Method}\\
\midrule
NASNet-A \cite{zoph2016neural}&2.65&3.3&1800&Single path&RL\\
AmoebaNet-B \cite{real2018regularized}&2.55$\pm$0.05 &\textbf{2.8}&3150&Single path&EA\\
Hierarchical Evo \cite{liu2017hierarchical}&3.75$\pm$0.12&15.7&300&Single path&EA\\
PNAS \cite{liu2018progressive}&3.41$\pm$0.09 &3.2&225&P Single path*&SMBO\\
IRLAS \cite{guo2018irlas}&2.60&3.91&-&Single path&RL\\
IRLAS-differential \cite{guo2018irlas}&2.71&3.43&-&Single path&RL\\
NAO* \cite{luo2018neural}&3.18&10.6&1000&Single path&gradient\\
NAO-WS \cite{luo2018neural}&3.53&\textbf{2.5}&-&Whole Supernet&gradient\\
ProxylessNAS \cite{cai2018proxylessnas}&\textbf{2.08}&5.7&-&Two path&gradient\\
\hline
SNAS \cite{xie2018snas}&2.85$\pm$0.02 &2.8&1.5&Whole Supernet&gradient\\
PARSEC \cite{casale2019probabilistic}&2.86$\pm$0.06 &3.6&0.6&Single path&gradient\\
GDAS \cite{GDAS}&2.93&3.4&\textbf{0.21}&Whole Supernet&gradient\\
BayesNAS \cite{zhou2019bayesnas}&2.81$\pm$0.04 &3.40$\pm$0.62&\textbf{0.2}&Whole Supernet&gradient\\
ENAS \cite{pham2018efficient}&2.89&4.6&-&Single path&RL\\
DARTS (1st) \cite{liu2018darts}&2.94&\textbf{2.9}&1.5&Whole Supernet&gradient\\
DARTS (2nd) \cite{liu2018darts}&2.76$\pm$0.09 &3.4&4&Whole Supernet&gradient\\
Random Search WS \cite{liu2018darts}&2.85$\pm$0.08 &4.3&2.7&Single path&random\\
\hline
EN$^2$AS &\textbf{2.64(2.59)}&3.1&\textbf{0.3}&Single path&novelty search\\
EN$^2$AS with performance reward &2.83&\textbf{2.3}&1&Single path&novelty search\\
EN$^2$AS with 1000 training epochs &\textbf{2.51(2.49)}&3.1&\textbf{0.3}&Single path&novelty search\\
\bottomrule
\end{tabular}
\caption{Comparison results with state-of-the-art NAS approaches on CIFAR-10. ``P Single path*" means Progressive Single path that the memory consumption progressively increases during the optimizing process. In ``NAO*", we only report the results of NAO with same number of initial channels. Our EN$^2$AS report the final test error (the best test error that this model have achieved).  We also report the result of 1000 training epochs of our searched best architecture.}
\label{table1}
\end{table*}

\section{Experiments and Results}

Experimental designs are following \cite{li2019random,liu2018darts,xie2018snas} for a fair comparison, which contain three stages: architecture search, architecture evaluation and transfer to larger datasets. We first perform our EN$^2$AS on small datasets, CIFAR-10 and PTB, to search for cell architectures on a smaller supernet architecture with fewer cells in the architecture search phase, then stack more multiple cells to construct larger architecture for full training and evaluation. Finally, the best-learned cells are also transferred to ImageNet and WikiText-2 to investigate the transferability \footnote{It is easy to reproduce our experiments results by replaceing cell structures in DARTS \cite{liu2018darts} with the structures shown in Fig.\ref{figure1}. Experimental results could be found in Appendix. We do not conduct experiments on hyperparameters tuning due to the computational resource constraint, and all hyperparameters could be found in Appendix.}.

\subsection{Architecture Search for Convolutional Cells}
The search space for convolutionary cells has been described in previous sections, and candidate operations are the same as DARTS which are also described in the Appendix. In the architecture search stage (first stage), the supernet is trained for 100 epochs with batch size 64 based on our novelty search based sampling method EN$^2$AS. After obtaining the most promising cell, we stack 20 cells for full training with batch size 96 for 600 epochs. The convolutional cell searched on CIFAR-10 is then transferred to ImageNet, following the mobile setting from \cite{liu2018darts}, and the other hyperparameters are also same as DARTS. These comparing approaches are divided into two groups: the first group approaches search on their own defined search space, and the search space for those approaches in the second group are the same as ours. Models for all approaches are trained with cutout.

\subsubsection{Results on CIFAR10}
The comparison results on CIFAR-10 with the state-of-the-art NAS methods are demonstrated in Table \ref{table1}. It is very impressive that the Random Search WS could obtain satisfactory results, which simply randomly sample architectures for supernet training. Random sampling strategy beats most One-Shot NAS, except DARTS (2nd) and BayesNet, with an elaborate controller in the same search space, which is also in line with the observation from \cite{bender2018understanding}. It is inspiring that the best architecture searched by our EN$^2$AS obtains the state-of-the-art test error on CIFAR-10 for weight sharing NAS with the same search space. Although ProxylessNAS performs better than ours, it searches on a different space that replaces all convolution layers in the residual blocks of a PyramidNet with tree-structured cells, and with more filters. Our approach is also very efficient that the architecture search phase only costs about 7.5 hours (0.3 days), and the memory consumption is the same as training a single architecture. The convolutional cell obtained by our EN$^2$AS is also very efficient, which has fewer parameters than most NAS methods. One thing we need to notice is that DARTS seems to conduct numerous experiments to find the best cell where its original version only gets $2.83\pm0.06$ test error, and our EN$^2$AS only conduct less than 10 experiments and find a cell structure that is better than DARTS.

\subsection{Architecture Search for Recurrent Cells}
We have also described the search space for recurrent cells in previous sections, and candidate operations are the same as DARTS which are also described in the Appendix. Both the embedding and the hidden sizes are set to 300 in the first stage which is the same as DARTS, and the supernet is trained using EN$^2$AS for 300 epochs with batch size 64. Then the embedding and the hidden sizes are changed to 850 for full training with 3600 epochs, and the best RNN cell is transferred to the WT2 dataset. These comparing approaches are also divided into two groups: the first group approaches are manually-designed, and the second group is based on NAS with the same search space as ours. GDAS is trained with 2000 epochs on PTB and 3000 epochs on WT2, and NAO is trained for 2000 epochs on the two datasets a to get results. The result of the rest approaches in the second group are reported in \cite{li2019random} which have the same number of training epochs with ours.

\subsubsection{Results on PTB}
The comparison results on PTB with the state-of-the-art manually-designed architectures and NAS methods are demonstrated in Table \ref{table2}. We can find that the DARTS achieve the state-of-the-art results on PTB among those NAS methods, which achieves a validation perplexity of 58.1 and a test perplexity of 55.7 and shows the efficiency of gradient method in the recurrent search space. NAO-ws also achieves an excellent results with with 56.7 test perplexity, while we need to notice that NAO set the $B=12$ in the RNN search space and our approach only use $B=8$ ordered nodes. Our EN$^2$AS obtains a competitive validation perplexity of 60.27 and a test perplexity of 57.8. Different from CNN results, the best manually-designed LSTM+15 SEs \cite{yang2017breaking} achieves excellent results with a validation perplexity of 58.1 and a test perplexity of 56.0, which is better than most NAS methods. One possible reason is that the search space of RNN is much simpler than CNN, and it is possible to find excellent architectures only relying on human experts. A further observation on our RNN experiments shows that all structures with excellent final results always perform very well in their early stages, which means that we could determine an architecture based on its performance on the proxy task (like using the validation performance with only 200 training epochs) with more certainty. This is also a possible reason why our EN$^2$AS performs worse than gradient-based methods on PTB, and with a simpler and less deceptive search space, these performance reward based search methods are more effective than our novelty based method. We further add the gradient of expected performance reward into adjusting the current architectures for architecture sampling (EN$^2$AS-PR), and the best-found architecture achieves a validation perplexity of 60.02 and a test perplexity of 57.36, which is on par with the state-of-the-art NAS methods on PTB.

\begin{table*} \footnotesize 
\centering

\begin{tabular}
{lcccccc}
\toprule

\makecell[c]{\multirow{2}*{Method}}&\multicolumn{2}{c}{Perplexity}&\makecell[c]{Parameters}&\makecell[c]{Search Cost}&\makecell[c]{\multirow{2}*{\makecell[c]{Memory\\Consumption}}}&\makecell[c]{\multirow{2}*{\makecell[c]{Search\\Method}}}\\
~&\makecell[c]{Valid}&\makecell[c]{Test}&\makecell[c]{(M)}&\makecell[c]{(GPU Days)}\\

\midrule
LSTM \cite{zoph2016neural}&60.7&58.8&24&-&-&manual\\
LSTM+SC \cite{merity2017regularizing}&60.9&58.3&24&-&-&manual\\
LSTM+15 SEs \cite{yang2017breaking}&58.1&56.0&22&-&-&manual\\

\hline
Random baseline\cite{xie2018snas}&64.1&61.5&23&-&Single path&random\\
NAS\cite{zoph2016neural}&-&64&25&1e$^{4}$&Single path&RL\\
ENAS \cite{li2019random}&60.8&58.6&24&0.5&Single path&RL\\
GDAS \cite{GDAS}&59.8&57.5&23&0.4&Whole Supernet&gradient\\
NAO-WS \cite{luo2018neural}&-&\textbf{56.6}&27&0.4&Whole Supernet&gradient\\
DARTS (1st) \cite{liu2018darts}&60.2&57.6&23&0.13&Whole Supernet&gradient\\
DARTS (2nd) \cite{liu2018darts}&\textbf{58.1}&\textbf{55.7}&23&0.25&Whole Supernet&gradient\\
Random Search WS \cite{liu2018darts}&57.8&55.5&23&0.25&Single path&random\\
Random Search WS* \cite{liu2018darts}&59.7&\textbf{57.16}&23&0.25&Single path&random\\

\hline
EN$^2$AS &60.27&57.8&23&0.17&Single path&novelty search\\
EN$^2$AS with performance reward &\textbf{60.02}&\textbf{57.36}&23&0.67&Single path&novelty\&reward\\

\bottomrule
\end{tabular}
\caption{Comparison results with state-of-the-art NAS approaches on PTB. In ``Random Search WS*", we retrain the best convolutional cell found by Random Search WS with the same hyperparameters setting like ours. In the EN$^2$AS with performance reward, we add the validation accuracy under the inherited weights into Eq.\ref{[8]}, and more details could be found in the Appendix}
\label{table2}
\end{table*}

\subsection{Discussion on the Combination of Novelty and Reward Search}
We further discuss the impact of adding a performance reward into adjusting architectures. Table \ref{table1} and Table \ref{table2} demonstrate the results of 4 different scenarios, where we conduct experiments with two different architecture update strategies (EN$^2$AS only depends on novelty as Eq.\eqref{[8]} and EN$^2$AS with performance reward (EN$^2$AS-PR) depends on the combination of novelty and reward Eq.\eqref{[9]}) on CIFAR-10 and PTB. We could find that there is no improvement in CNN when combining with performance reward in our EN$^2$AS, while it could greatly enhance the performance of our EN$^2$AS on RNN. Adjusting the trade-off between novelty and reward may be a solution to improve the performance of EN$^2$AS-PR, while the hyperparameter tuning is not the scope of this paper due to the computational resource constraint. More interesting, during the evaluation of the architectures in RNN experiments, we find that most architectures obtained by EN$^2$AS-PR outperforms the best architecture searched by original EN$^2$AS. The possible reason maybe that the search space of RNN is less complicated and deceptive than CNN. Furthermore, as described in previous experiments, the performance of early stages in RNN is very informative, which means we could determine an architecture based on its performance with inherited weights with more certainty. However, this combination of novelty and performance reward introduces inference into the supernet training, and it needs to evaluate numerous architectures that makes it not as efficient as our original EN$^2$AS. With the same experimental settings, EN$^2$AS only cost 0.3 and 0.17 GPU day on CIFAR-10 and PTB, while EN$^2$AS-PR cost more than 1 and 0.67 GPU day, respectively.

\subsection{Discussion on Architecture Sampling and Model Selection}
In the One-Shot NAS, it usually contains two important stages: architecture sampling for supernet training, and model selection from the trained supernet. In the architecture sampling stage, we consider three different methods, random sampling \cite{li2019random}, our novelty search, and also novelty with reward-based sampling. And in the model selection stage, we also consider two different methods, random search (RS) and evolutionary algorithm (EA). Table \ref{table3} demonstrates the results of 6 scenarios, where we conduct experiments on CIFAR-10 and PTB with different architecture sampling and model selection methods, respectively. During this experiment, Random sampling + RS is the same as Random Search WS, while we are not able to achieve results as excellent as Random Search WS with less than 10 independent experiments. The first thing that we could find from the table is that the combination of novelty search and evolutionary algorithm obtains the best results on CIFAR-10, and the combination of novelty with reward and evolutionary algorithm achieve the best results both in CIFAR-10, which shows the superiority of EA than random search in model selection. Furthermore, we could observe from this table that, with the same supernet training method, the evolutionary algorithm clearly outperforms random search in most cases. However, with the same model selection method, novelty search is not guaranteed to obtain better results than random sampling. These results also show the importance of supernet training that specific supernet training strategy may need to be devised for different tasks in the neural architecture search.

\begin{table} \footnotesize
\centering

\begin{tabular}
{lccc}
\toprule
\makecell[c]{\multirow{2}*{Method}}&\multicolumn{1}{c}{CIFAR-10}&\multicolumn{2}{c}{PTB}\\
~&\makecell[c]{Test(\%)}&\makecell[c]{Valid}&\makecell[c]{Test}\\

\midrule
Random sampling + RS &3.01&60.73&58.20\\
Random sampling + EA &2.98&60.42&57.88\\
Novelty Search + RS &2.95&60.46&58.33\\
Novelty Search + EA &\textbf{2.64}&60.27&57.82\\
Novelty\&Reward + RS &3.05&60.22&58.03\\
Novelty\&Reward + EA &2.83&\textbf{60.02}&\textbf{57.36}\\

\bottomrule
\end{tabular}
\caption{Comparison results with different architecture sampling and architecture selection methods.}
\label{table3}
\end{table}

\section{Conclusion and future work}
This paper originally focuses on how to make the supernet more predictive for weight-sharing neural architecture search and proposes a novelty search based controller which samples architectures based on novelty to train the supernet. In particular, a novelty search mechanism is developed to efficiently find the most abnormal architecture, and the single-path model is adopted to greatly reduce computational and memory demand. Experimental results demonstrate the superiority of our approach which could find the state-of-the-art or competitive CNN and RNN models, which suggest that our approach makes the supernet much more predictive than other NAS methods. In our future work, we focus on leveraging human knowledge in neural architecture search to enhance its transferable ability. Furthermore, how to transform the discrete architecture space into a continuous space is also one of our future work directions.

%\bibliographystyle{aaai}
%\bibliography{aaai2020}

\section{Appendix}
\subsection{Experimental details}
As discussed in the previous, neural architecture search generally contains three stages: architecture search, architecture evaluation and transfer to larger datasets, and all experimental settings are following DARTS in this paper for fair comparison.

\subsubsection{Searching for Convolutionary cells}
The search space is following \cite{liu2018darts,pham2018efficient,li2019random}, which contains 7 different operation: $3\times 3$ separable convolution, $5\times 5$ separable convolution, $3\times 3$ dilated separable convolutions, $5\times 5$ dilated separable convolutions, $3\times 3$ max pooling, $3\times 3$ average pooling, identity, and $zero$, where $zero$ means there is no operation which helps to compress the neural network. The convolutional operations use ReLU-Conv-BN order, and separable convolutions applies ReLU-Conv-BN for twice. The covolutional cell contains 7 nodes: two input nodes, 4 operation nodes and 1 output node.  There are two types of cells in our CNN: normal cell and reduction cell, encode as $(\alpha_{normal},\alpha_{reduce})$. Reduction cells only locate in the 1/3 and 2/3 of the total depth of our network, and the operations adjacent to the input nodes in reduction cells are of stride two. 

We first stack 8 convolutionary cells to build the architecture for architecture search, where the number of initial channels $c$ is set as 16, the initial SGD learning rate is 0.025 and annealed down to 0.001 with a cosine schedule, the cutout length is 16, path dropout probability is 0.4, momentum 0.9, and weight decay $3\times 10^{-4}$. The supernet is trained for 100 epochs with batch size 64 to get the most promising cell and we divide the training dataset of CIFAR-10 into two half as training and validation dataset in the architecture search stage. We then stack 20 cells for full training with batch size 96 for 600 epochs, where the initial channel is increase to 36, auxiliary towers with weight 0.4, path dropout probability is set as 0.2, and other hyperparameters remain the same. The best covolutional cell searched on CIFAR-10 is then transferred to ImageNet. We also follow the mobile setting from \cite{li2019random,liu2018darts,xie2018snas} with 224$\times$224 input image size and the number of multiply-add operations is restricted to be less than 600M, weight decay is $3\times 10^{-5}$, and initial SGD learning rate is 0.1 with decayed factor of 0.97. The network is stacked by 14 cells with batch size 128 with 250 epochs training.

\subsubsection{Searching for Recurrent cells}

The search space for RNN is also following \cite{liu2018darts,pham2018efficient,li2019random}, which contains 4 different operation: $tanh$, $relu$, $sigmoid$, $identity$. The recurrent cell contaions 12 nodes: two input nodes, 1 adding nodes, 8 operation nodes and 1 output node, where the adding node is to add two inputs and apply $tanh$ activation function. The input of each node is output of one of it previous nodes after applied operation and the hidden state $h_t$ is calculated based on the input $x_t$ and its former hidden state $h_{t-1}$, and the output of the cell is the summation of outputs of all operation nodes, and we need only encode one type cell in RNN architecture.

In the recurrent architecture search, both the embedding and the hidden sizes are set to 300, we use a SGD optimizer with learning rate of 20.0, BPTT length 35, and weight decay $5\times 10^{-7}$, 0.2 dropout rate for word embeddings, 0.75 for the cell input, 0.25 for all the hidden nodes, and 0.75 to output layer. The supernet is trained for 300 epochs with batch size 64 to get the most promising cell. Then the embedding and the hidden sizes are changed to 850 for full training  with  3600  epochs with 64 batch size for the best found recurrent cell, where we use a averaged SGD optimizer with 20 initial learning rate and weight decay $8\times 10^{-7}$, the token-wise dropout on the embedding layer is set to 0.1, and other hyperparameters are same as before. The best RNN cell is then transferred to WT2 dataset, where the embedding and hidden sizes are changed to 700, weight decay to $5\times 10^{-7}$, and hidden-node variational dropout to 0.15. 

\subsubsection{Hyperparameters setting for novelty calculation}
As described in the previous, the gradient of expected novelty is calculated based on Eq.\eqref{[7]}, where we set $n=10$, $\sigma=1$. And we update the architectures in the archive based on Eq.\eqref{[8]}, where we set $\gamma=0.1$. We calculate the novelty between the sampled architecture $\alpha$ and the archive $A$ based on Eq.\eqref{[6]}, which is calculated as the mean distance of $\alpha$ and its k-nearest neighbors from $A$, where $k=10$ and the archive size $size(A)=100$. We set these hyperparameters for novelty calculation all the same in all experiments. As to the the calculation of distance between architectures, we could individually compare the difference of input edges of each node because the order of nodes is fixed, where the two edges for the same node in two architectures are seen as same only when the input node and the operation applied to it are same. 

\begin{figure*}
\centering

  \subfloat[Normal cell learned on CIFAR-10]{
  \begin{minipage}{6cm}
      \includegraphics[width=6cm,height=3.5cm]{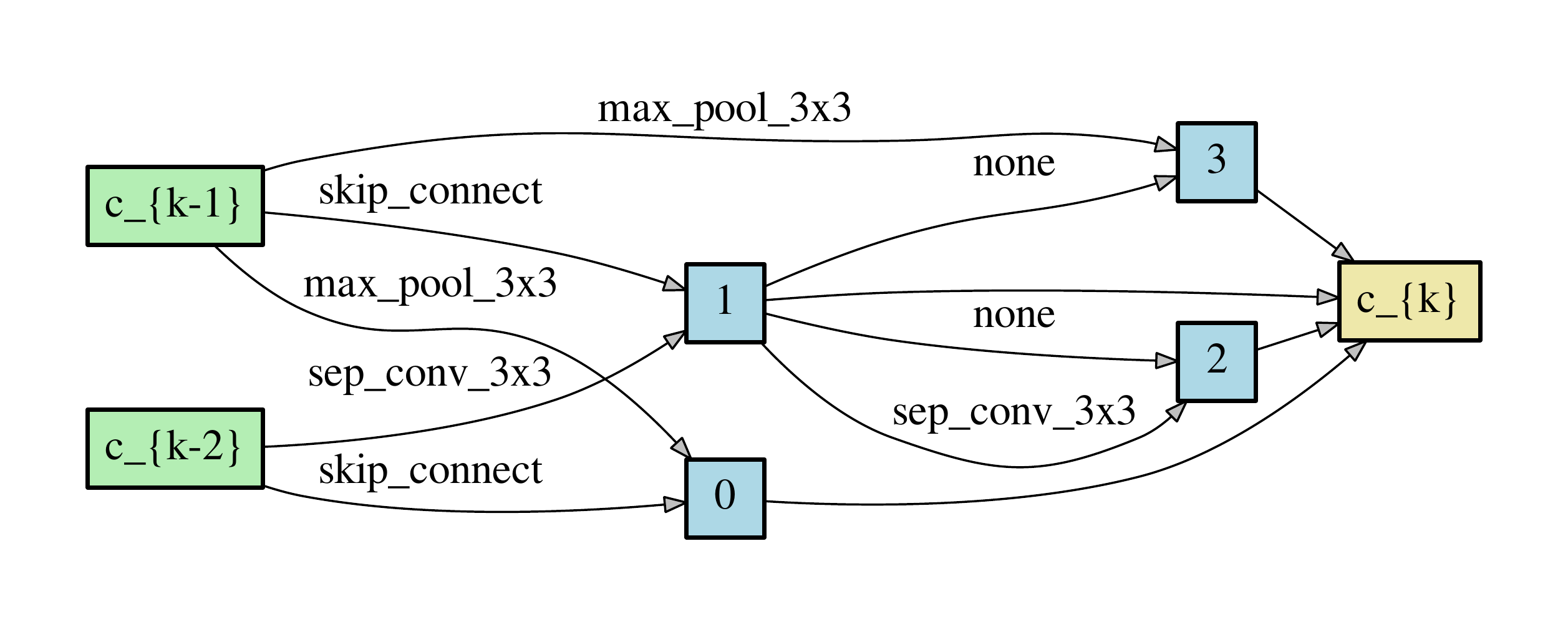}
  \end{minipage} 
  }
   \subfloat[Reduction cell learned on CIFAR-10]{
  \begin{minipage}{6cm}
      \includegraphics[width=6cm,height=3.5cm]{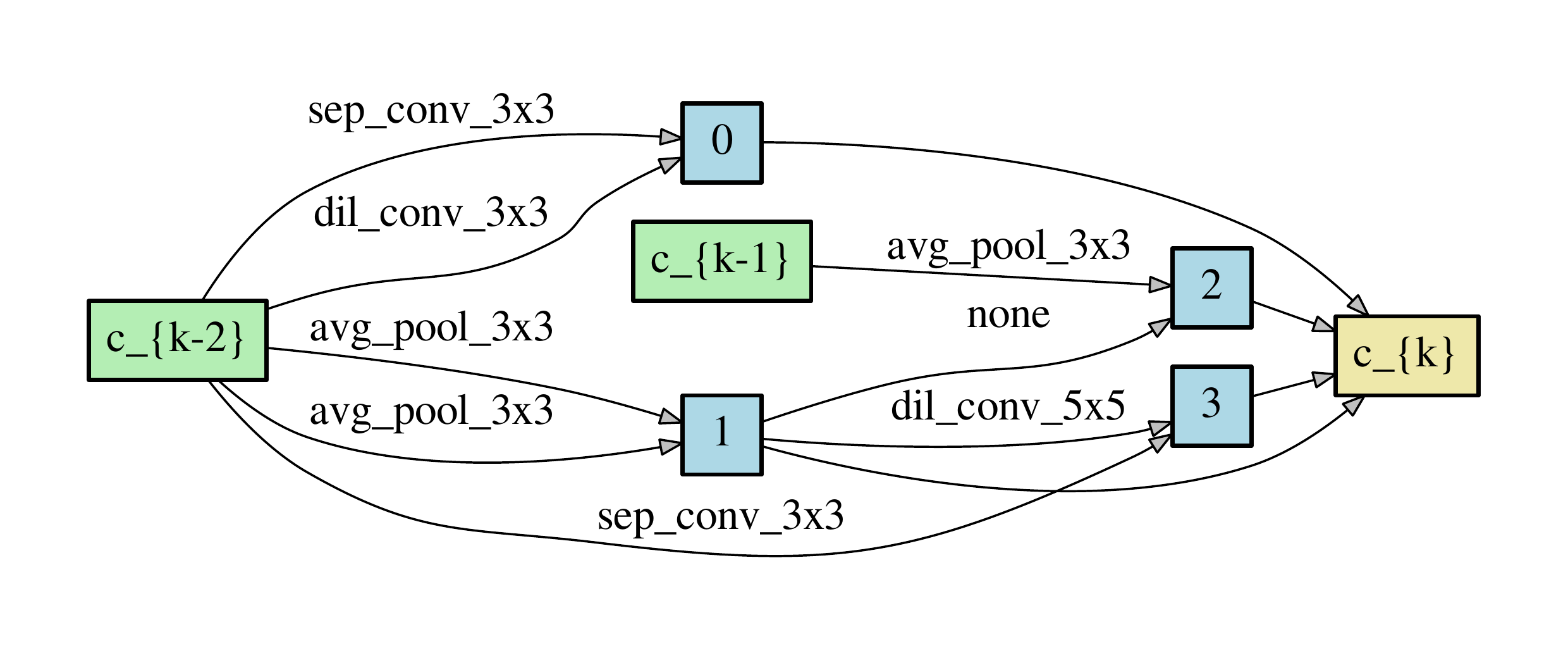}
  \end{minipage}
  }
   \subfloat[Recurrent cell learned on PTB]{
  \begin{minipage}{6cm}
      \includegraphics[width=6cm,height=3.5cm]{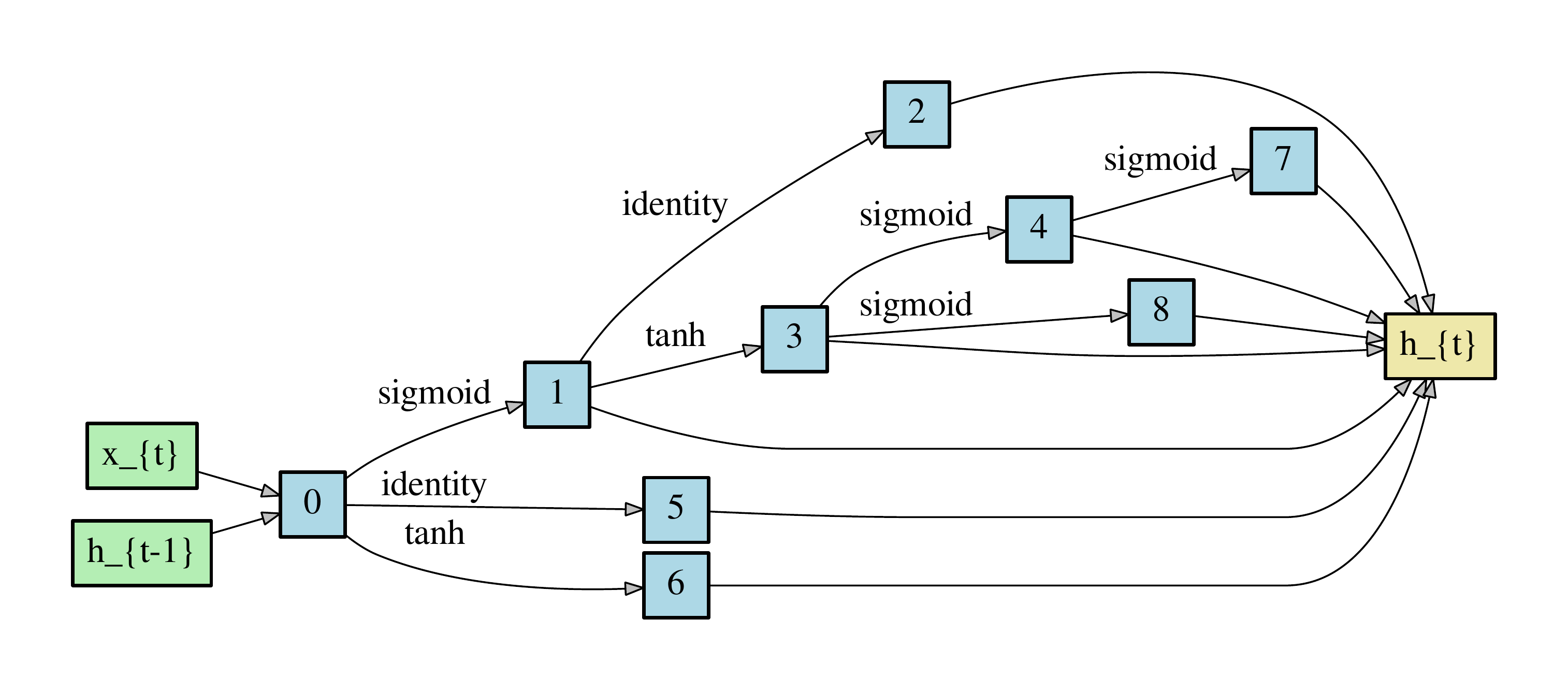}
  \end{minipage}
  }
   \caption{Other excellent cell structures found by our algorithms. The CNN cell structure in this figure is obtained by EN$^2$AS with performance reward, and the RNN cell structure in this figure is obtained by EN$^2$AS}
   \label{figure2}
\end{figure*}

\subsection{Algorithms implementation}

Our algorithm is based on DARTS \cite{liu2018darts}, which makes our algorithm very easy to be implemented. DARTS searches an architecture weight $w$ based on gradient method in each search epoch, which is updated with all weights of supernet. The solution of our EN$^2$AS is also encoded as an architecture weight $w$, while it only contains $0$ and $1$, and the supernet will not update the weights of those connections corresponding to architecture weight 0, and therefore the memory consumption for our supernet training is same as training a single path architecture. The novelty-driven controller of our EN$^2$AS is to generate an architecture in each epochs based on equation \eqref{[8]}, and the corresponding architecture weights are set as 1 and other as 0, and then we also follow DARTS to train the supernet. After training the supernet, we apply a baseline evolutionary algorithm on the trained supernet with validation dataset $\mathbb{D}_{val}$ to find the most promising architecture, where  we measure each architecture based on Eq.\eqref{[4]}.

\subsubsection{Results on ImageNet}
The comparison results on ImageNet of all state-of-the-art NAS are present in Table\ref{table4}. All NAS methods transfer the searched cell architecture on CIFAR-10 to ImageNet, only except ProxylessNAS which searches on ImageNet. Our model could obtain competitive result with Top1/Top5 test error 27.08\%/8.88\% with only 4.5M parameters. We could observe that our model beats all approaches with same search space with the lowest Top5 test error and competitive Top1 test error, except PARSEC which trains its model on ImageNet for 600 epochs while we only train our model for 250 epochs. We could also find that IRLAS \cite{guo2018irlas} obtains the best performance when it is transferred to ImageNet, and it even beats ProxylessNAS which directly search on ImageNet. The reason that IRLAS perform so excellent on large dataset maybe that it takes the human knowledge into searching architectures and makes the searched cell have simple structure.

\begin{table} \footnotesize
\centering

\begin{tabular}
{p{120pt}{l}p{20pt}{c}p{15pt}{c}p{20pt}{c}}
\toprule
\makecell[c]{\multirow{2}*{Method}}&\multicolumn{2}{c}{Test Error}&\makecell[c]{Parameters}\\
~&\makecell[c]{Top1}&\makecell[c]{Top5}&\makecell[c]{(M)}\\

\midrule

NASNet-A \cite{zoph2016neural}&26.0&8.4&5.3\\
AmoebaNet-B \cite{real2018regularized}&25.5&8.0&5.3\\
PNAS \cite{liu2018progressive}&25.8&8.1&5.1\\
IRLAS-mobile \cite{guo2018irlas}&24.72&-&-\\
ProxylessNAS \cite{cai2018proxylessnas}&24.9&7.5&-\\

\hline

SNAS \cite{xie2018snas}&27.3&9.25&4.3\\
PARSEC* \cite{casale2019probabilistic}&26.3&8.4&5.5\\
GDAS* \cite{GDAS}&27.5&9.1&4.4\\
BayesNAS \cite{zhou2019bayesnas}&26.5&8.9&3.9\\
DARTS \cite{liu2018darts}&26.0&9.0&4.9\\

\hline
EN$^2$AS &27.08&8.88&4.5\\

\bottomrule
\end{tabular}
\caption{Comparison results with state-of-the-art NAS approaches on ImageNet. ``PARSEC*" follows the same hyperparameters setting as us while the architecture is trained for 600 epochs. In ``GDAS*", we only report results of GDAS that satisfy ImageNet-mobile setting.}
\label{table4}
\end{table}

\subsubsection{Results on WT2}
Promising models on PTB obtained by different NAS methods are then transferred to WT2. The results of different models on WT2 are presented on Table\ref{table5}. We can find that the manually-designed models are supposed to achieve better performance than NAS methods. This phenomenon shows that the transferable ability of the discovered models on PTB are a little bit weak, and designing simple structures and taking the human knowledge into automatically searching architectures, like IRLAS\cite{guo2018irlas}, maybe beneficial to search for models with better transferable ability.

\begin{table} \footnotesize
\centering

\begin{tabular}
{p{120pt}{l}p{20pt}{c}p{15pt}{c}p{20pt}{c}}
\toprule
\makecell[c]{\multirow{2}*{Method}}&\multicolumn{2}{c}{Test Error}&\makecell[c]{Parameters}\\
~&\makecell[c]{Valid}&\makecell[c]{Test}&\makecell[c]{(M)}\\

\midrule

LSTM \cite{zoph2016neural}&69.1&65.9&33\\
LSTM+SC \cite{merity2017regularizing}&69.1&65.9&23\\
LSTM+15 SEs \cite{yang2017breaking}&66.0&63.3&33\\

\hline
ENAS \cite{xie2018snas}&72.4&70.4&33\\
GDAS \cite{GDAS}&71.0&69.4&33\\
NAO \cite{luo2018neural}&-&67.0&36\\
DARTS \cite{liu2018darts}&71.2&69.6&33\\
\hline
EN$^2$AS &73.90&71.56&33\\
\bottomrule
\end{tabular}
\caption{Comparison results with state-of-the-art NAS approaches on WT2.}
\label{table5}
\end{table}

\end{document}